\documentclass[english]{cccconf}
\usepackage[comma,numbers,square,sort&compress]{natbib}
\usepackage{epstopdf}

\usepackage{amsmath}
\usepackage[noend]{algpseudocode}

\usepackage{algorithmicx,algorithm}
\begin{document}

\title{Spectrum Attention Mechanism for Time Series Classification}

\affiliation[zhou]{Institute of Cyber-Systems and Control, College of Control Science and Engineering, Zhejiang University, Hangzhou, 310027, China
        \email{zhoushibo@zju.edu.cn}}
\affiliation[pan]{Institute of Cyber-Systems and Control, College of Control Science and Engineering, Zhejiang University, Hangzhou, 310027, China
	\email{ypan@zju.edu.cn}}

\author{Shibo Zhou\aref{zhou},
Yu Pan\aref{pan}}

\maketitle

\begin{abstract}
Time series classification(TSC) has always been an important and challenging research task. 
With the wide application of deep learning, more and more researchers use deep learning models to solve TSC problems. Since time series always contains a lot of noise, which has a negative impact on network training, people usually filter the original data before training the network. The existing schemes are to treat the filtering and training as two stages, and the design of the filter requires expert experience, which increases the design difficulty of the algorithm and is not universal. We note that the essence of filtering is to filter out the insignificant frequency components and highlight the important ones, which is similar to the attention mechanism. In this paper, we propose an attention mechanism that acts on spectrum (SAM). The network can assign appropriate weights to each frequency component to achieve adaptive filtering. We use L1 regularization to further enhance the frequency screening capability of SAM. We also propose a segmented-SAM (SSAM) to avoid the loss of time domain information caused by using the spectrum of the whole sequence. In which, a tumbling window is introduced to segment the original data. Then SAM is applied to each segment to generate new features. We propose a heuristic strategy to search for the appropriate number of segments. Experimental results show that SSAM can produce better feature representations, make the network converge faster, and improve the robustness and classification accuracy.

\end{abstract}


\footnotetext{This work was supported by the National Key 
	R\&D Program of China under Grant 2018YFB1700100.}

\section{Introduction}

Time series are real-valued ordered data with the characteristics of large data volume, high dimensionality, and high noise. The traditional TSC algorithm requires manual feature extraction, which is complicated and not universal\cite{1}. With the wide application of deep learning in computer vision, natural language processing, recommendation system, etc.\cite{2,3,4}, more and more researchers use deep learning to solve TSC problems. Deep learning based TSC model do not require manual design of features, the network can automatically learn the internal patterns of data through training. Typical TSC models include FCN\cite{5}, MCNN\cite{6}, InceptionTime\cite{7}, etc. However, time series usually contain a lot of noise, which has a negative impact on the training of the model. Before applying the algorithm, people generally filter the original data to improve the feature representation. The existing schemes are to treat the filtering and classifying as two stages. For example, the effective frequency spectrum of EMG signal ranges from 0 to 4hz, digital filters are generally applied to filter out high-frequency noise in the data preprocessing stage\cite{8}. For EEG and MEG time series, a high-pass filter is generally used to remove slow drifts\cite{9}. In the field of remote sensing images, people usually apply the fourier smoothing algorithm to the original normalized difference vegetation index time series (NVDI-TS) to improve the classification accuracy\cite{10}. In all the above cases, the filter design requires expert experience, which undoubtedly increases the difficulty of algorithm design.

In this paper, we propose a spectrum attention mechanism (SAM) that is compatible with the deep learning model. It is embedded in the first layer of the network, and the mask vector is updated through training, so as to achieve adaptive filtering of the original data and generate the features that is more conducive to network training. We validate the effectiveness of the scheme through experiments on synthetic datasets and real datasets.

The paper is organized as follows: In Section 2, the problem formulation and our methods are presented. Section 3 presents the experiments and results. Finally, Section 4 provides the main conclusions of the paper.

\section{Methodology}
\subsection{Frequency Domain Filtering}
Our goal is to design an adaptive filtering module that is compatible with the deep learning models, which can generate a feature representation that is more conducive to network training. We notice that the common point of existing schemes is the use of frequency domain filtering. That is, the frequency domain transformation of the original data is carried out to get the spectrum, and then the specific frequency components are filtered out according to specific scenes. This process can be described by the formula~(\ref{e1}).
\begin{equation}\label{e1}
x_{{filtered }}^{n}=\mathcal{F}^{-1}\left(f\left(\mathcal{F}\left(x^{n}\right), { mask }\right)\right)
\end{equation}

\noindent Where $\mathcal{F}$ denotes the frequency domain transformation, $mask$ is a vector of the same dimension as the input data, and $f$ represents the operation on the spectrum, which is generally a dot product. 

We should preserve the important frequency components in the spectrum and remove the unimportant components, so the key to the problem is how to determine a suitable mask.

\subsection{Discrete Cosine Transformation}
We use the Discrete Cosine Transform (DCT) to transform the raw data into the frequency domain. DCT is a Fourier-related transform similar to the discrete Fourier transform (DFT), but uses only cosine basis. DCT for a time series $X$ of length $N$ is defined as formula~(\ref{e2}) and~(\ref{e3}).
\begin{equation}\label{e2}
\begin{aligned}
&X[k]=\sum_{n=0}^{N-1}a(k) x[n] cos \left(\frac{(2 n+1) \pi k}{2 N}\right)&
\end{aligned}
\end{equation}

\begin{equation}\label{e3}
\begin{aligned}
&X[k]=\sum_{n=0}^{N-1} a(n) x[n] cos \left(\frac{(2 k+1) \pi n}{2 N}\right)&
\end{aligned}
\end{equation}

where $a(u)=\left\{\begin{array}{l}\sqrt{\frac{1}{N}}, u=0 \\ \sqrt{\frac{2}{N}}, u=1,2, \cdots, N-1\end{array}\right.$

We choose to use DCT because it has the following advantages compared with DFT:
\begin{itemize}
	\item DCT coefficients are real numbers as opposed to the DFT complex coefficients, which is more suitable for gradient descent.
	\item DCT can handle signals with trends well, while DFT suffers from the problem of "frequency leakage" when representing simple trends.
	\item When the successive values are highly correlated, DCT achieves better energy concentration than DFT.
\end{itemize}

\subsection{Spectrum Attention mechanism (SAM)}
In cognitive science, due to the bottleneck of information processing capabilities, humans selectively focus on part of the information while ignoring the rest of the visible information. This is usually called the attention mechanism\cite{11}. Since frequency domain filtering is to remove unimportant frequency components and retain or strengthen important components, this is the same idea as the attention mechanism.
\begin{figure}[!htb]
  \centering
  \includegraphics[width=\hsize]{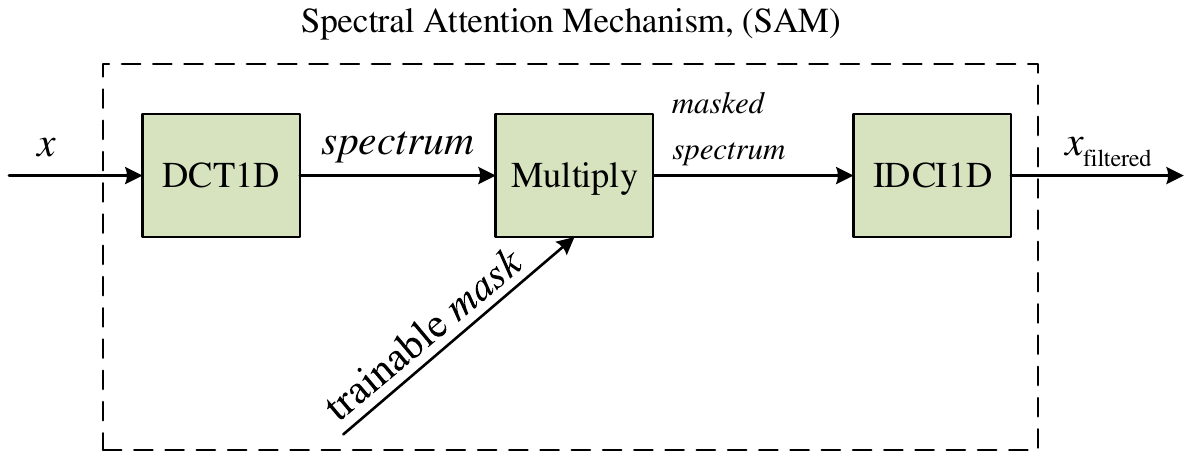}
  \caption{Spectral Attention Mechanism}
  \label{f1}
\end{figure}

We design an attention mechanism that acts on the spectrum (see Fig.~\ref{f1}). It contains trainable parameters $mask$ of the same dimension as the input signal, representing the weight of each frequency component, and the initial value is 1. The weights are updated through training, so as to realize adaptive filtering and generate better features. The forward propagation of this layer is summarized in Algorithm~\ref{algo1}.

\begin{algorithm}[h]
\caption{Spectral Attention Mechanism(SAM)} 
\label{algo1}
\hspace*{0.02in} {\bf Input:} 
Univariate time series ${{x}^{n}}$\\
\hspace*{0.02in} {\bf Output:} 
$x_{filtered}^{n}$\\
\hspace*{0.02in} {\bf Initialization:}
All-ones learnable array $mask^{n}$
\begin{algorithmic}[1]
\State \# Transform the input series into spectral domain

\noindent$s{{p}^{n}}\leftarrow DCT({{x}^{n}})$

\State \# Element-wise multiply $spectrum$ by $mask$

\noindent$masked\_s{{p}^{n}}\leftarrow s{{p}^{n}}\cdot mas{{k}^{n}}$ 

\State \# Transform the $spectrum$ back into the time domain

\noindent${{x}_{filtered}}\leftarrow IDCT(masked\_s{{p}^{n}})$ 
\end{algorithmic}
\end{algorithm}


\begin{figure}[!htb]
  \centering
  \includegraphics[width=\hsize]{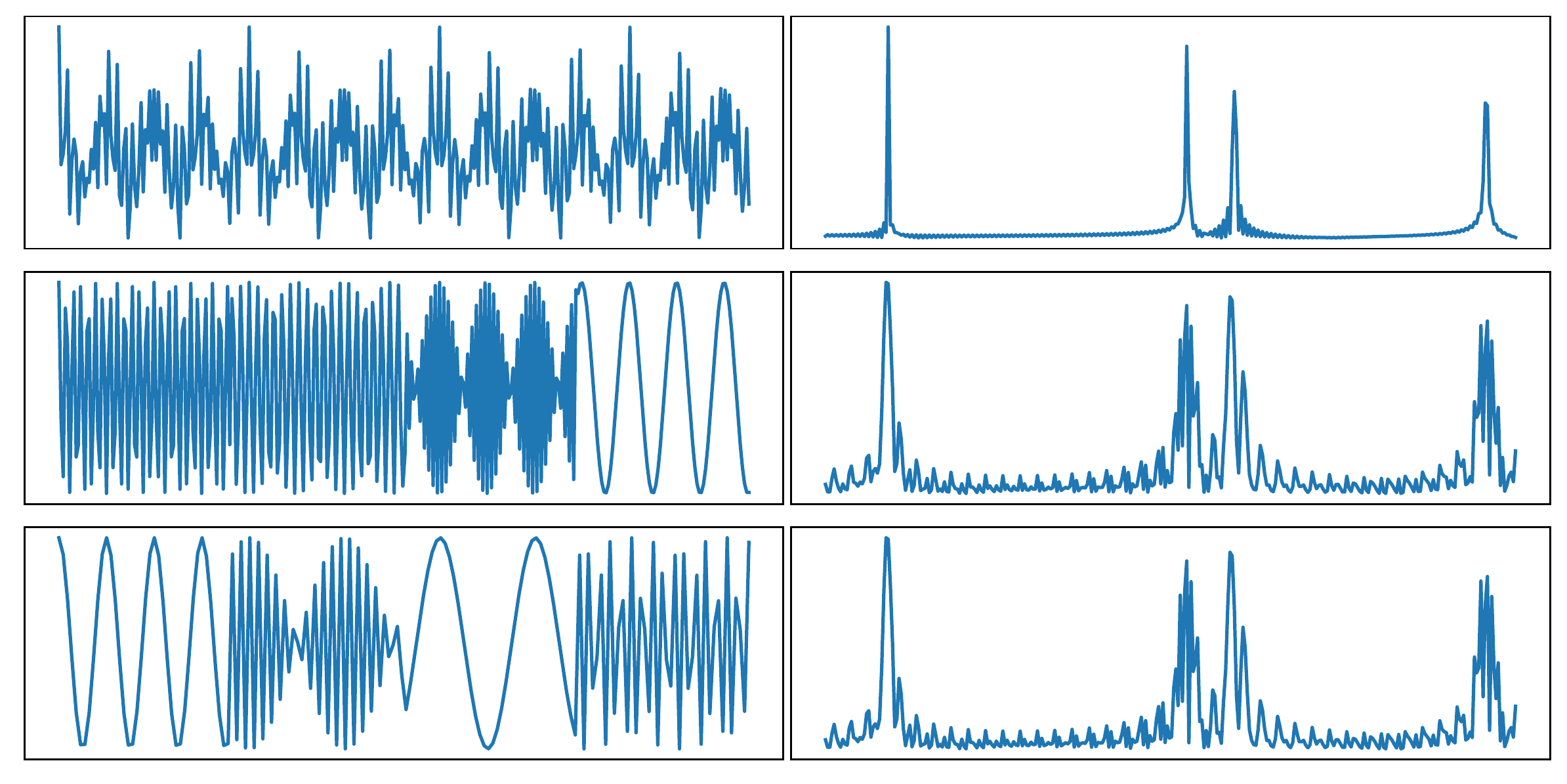}
  \caption{Three time series (left) and their corresponding spectrum (right).}
  \label{f2}
\end{figure}
SAM uses the spectrum of the entire sequence, which cannot reflect the phase information of the original signal. As shown in Fig.~\ref{f2}, although the three time series have great differences in the time domain, their spectrum is very similar. This is because they contain the same frequency components, only the phase of each frequency component is different. Therefore, SAM has inherent defects in processing non-stationary signals. However, almost all signals are non-stationary in real world. In order to retain part of the valuable phase information, we use a tumbling window to divide the original sequence into $K$ segments of equal length, and apply SAM to each segment. The SAM output of each segment is concatenated on the channel dimension as output features. The main algorithm is summarized in Algorithm ~\ref{algo2}.

\begin{algorithm}[h]
\caption{Segmented SAM (SSAM)} 
\hspace*{0.02in} {\bf Input:} 
${{x}^{n}}$, number of segments $K$\\
\hspace*{0.02in} {\bf Output:} 
generated features $x_{new}^{T \times K}$
\label{algo2}
\begin{algorithmic}[1]
\State $T\leftarrow n//K$ \# Initialize length of each segment

\State ${{x}_{new}}\leftarrow zeros(T,K)$ \# Initialize ${{x}_{new}}$

\For{ i = 1 to $K$ }

\State \# Get the ${{i}^{th}}$ segment

\indent$cur\leftarrow x[(i-1)*T:i*T]$

\State \# Apply SAM to the ${{i}^{th}}$ segment\

$cur\_output\leftarrow SAM(cur)$

\State \# Update output

${{x}_{new}}[:,i]\leftarrow cur\_output$

\EndFor
\State \textbf{end for }
\end{algorithmic}
\end{algorithm}
\begin{algorithm}[h]
\caption{Searching for the best number of segments} 
\hspace*{0.02in} {\bf Input:} 
training dataset, validation dataset\\
\hspace*{0.02in} {\bf Output:} 
\label{algo3}
${K}_{best}$
\begin{algorithmic}[1]
\State $min\_loss\leftarrow Inf$

\State ${{K}_{best}}\leftarrow None$

\For{  K = 1 to 8  }

\State Train network for 5 epochs

\State  Calculate the validation loss

\If{validation loss $<$ min\_loss}

\State    ${{K}_{best}}\leftarrow K$ 

\State    $min\_loss\leftarrow validation\_loss$

\EndIf
\State \textbf{end if }
\EndFor
\State \textbf{end for }
\end{algorithmic}
\end{algorithm}
Since $K$ is a hyperparameter, which difficult to determine directly. We design a heuristic search method, as shown in algorithm~\ref{algo3}. The candidate range of $K$ is set to 1 to 10, that is, the original data is divided into 10 segments at most. Take each $K$ value, train the model for 5 epochs, and apply the $K$ corresponding to the minimum validation loss to the final model.

\subsection{Architecture}
\begin{figure}[!htb]
  \centering
  \includegraphics[width=\hsize]{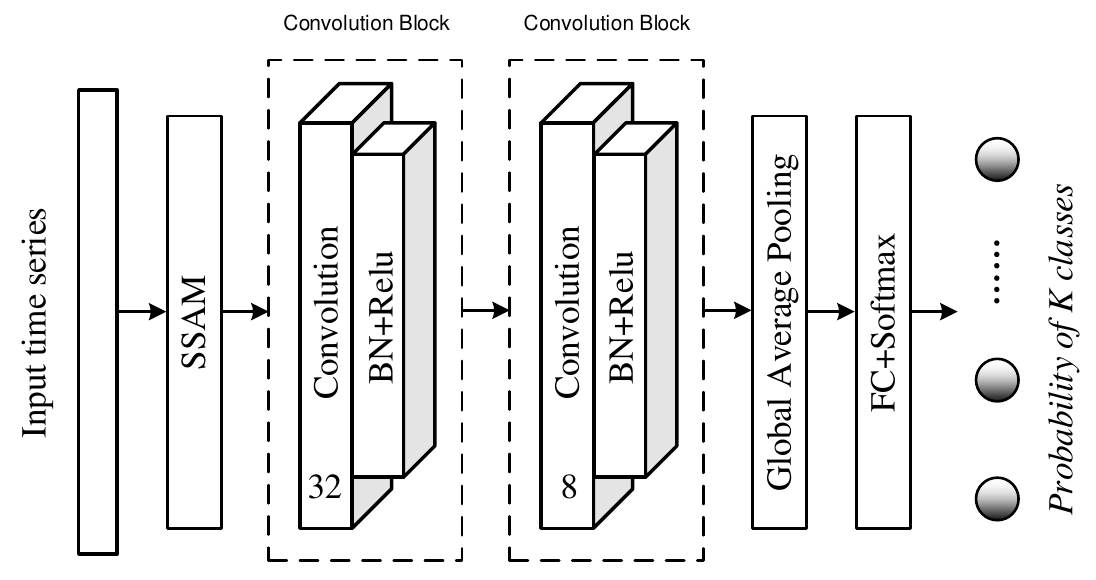}
  \caption{The model architecture of SSAM-CNN.}
  \label{f3}
\end{figure}
In order to avoid the strong fitting ability of complex models to cover up the performance of SSAM, wo present a relatively simple model. We first define the convolution block, which consists of one-dimensional convolutional layer, batch normalization layer\cite{12}, and activation layer. The basic convolution block is:
\begin{equation}\label{e5}
\begin{array}{l}
y=\omega \otimes x+b \\
s=B N(y) \\
h={relu}(s)
\end{array}
\end{equation}

$ \otimes $ is the convolution operator. Our model is shown in Figure~\ref{f3}. First, the raw data is input to SSAM for filtering to generate features more suitable for network training. Then input into two convolution blocks to extract features. The kernel size is \{8,5\}, and the channel dimension is \{32,8\}. After the convolution blocks, the features are fed into a global average pooling layer\cite{13} instead of a fully connected layer, which largely reduces the number of weights. The final label is produced by a softmax layer. 

\section{Experiment}
In this section, experiments will be conducted on a synthetic dataset and four widely used real datasets from UCR archive\cite{14}. It should be noted that our goal is to verify the effectiveness and universality of the design, so we did not perform an overly detailed search on the hyperparameters of the model.
\subsection{Data}
\noindent
\textbf{Synthetic}: To better understand the relationship between model performance and the characteristics of the data, we define 3 classes: $C1$, $C2$, and $C3$ and generate 2000 series from each as follows:
\begin{equation}\label{e6}
x_{t}=\cos \left(\frac{2 \pi t}{100}\right)+cos \left(\frac{2 \pi 5 t}{100}\right)+\omega_{t} \text { for } x_{t} \in C_{1}
\end{equation}
\begin{equation}\label{e7}
x_{t}=\cos \left(\frac{2 \pi t}{100}\right)+cos \left(\frac{2 \pi 20 t}{100}\right)+\omega_{t} \text { for } x_{t} \in C_{2}
\end{equation}
\begin{equation}\label{e8}
x_{t}=\cos \left(\frac{2 \pi t}{100}\right)+cos \left(\frac{2 \pi 80 t}{100}\right)+\omega_{t} \text { for } x_{t} \in C_{3}
\end{equation}
where $t = 1,..,100$,and $\omega_{t}$ is a gaussian noise with standard deviation $\sigma=2$. 

The most dominant frequencies for $C_{1}$ are 1 and 5, while for $C_{2}$ are 1 and 20, while for $C_{3}$ are 1 and 80. Due to the influence of noise, the series from the three classes look similar in the time domain.\\
\textbf{CBF}: This is a shape classification datasets which contains three classes: Cylinder, Bell and Funnel.\\
\textbf{Control Charts (CC)} : This dataset is derived from the control chart, which contains six different control modes: normal, cyclic, increasing trend, decreasing trend, upward shift and downward shift.\\
\textbf{Face}: This dataset originates from a face recognition problem. It consists of four different individuals, making different facial expressions. The task is to identify the person based on the head profile, which is represented as ``pseudo time series''.\\
\textbf{Trace}: This dataset records instrument data from a nuclear power plant for fault detection.

The details of each dataset is summarized in Table~\ref{t1}.

\begin{table}[!htb]
  \centering
  \caption{The characteristics of each dataset}
  \label{t1}
  \begin{tabular}{c|c|c|c}
    \hhline
Datasets& Classes& Instances& Time Series length\\\hline
 Synthetic&	3&	6000&	100\\\hline
 CBF&	3&	310&	128\\\hline
  CC&	6&	600&    60\\\hline
Face&	4&	1120&	350\\\hline
Trace&	4&	2000&	275\\\hline

    \hhline
  \end{tabular}
\end{table}

\subsection{Experimental settings}
We first normalize the data, and then divide it into training dataset, validation dataset, and test dataset at a ratio of 6:2:2. Then algorithm 3 is applied to search for the optimal number of segments. In the training stage, we record the validation loss in each epoch, and select the model with the minimum validation loss as the final model. Some of the hyperparameter configurations are shown in Table~\ref{t2}.

\newcommand{\tabincell}[2]{\begin{tabular}{@{}#1@{}}#2\end{tabular}}  
\begin{table}[!htb]
  \centering
  \caption{Hyperparameters of SSAM-CNN model.}
  \label{t2}
  \begin{tabular}{c|c|c|c|c}
    \hhline
\tabincell{c}{learning\\rate} & \tabincell{c}{learning\\algorithm} & \tabincell{c}{regularization\\coefficient} &epochs&	\tabincell{c}{batch\\size}\\
\hline
0.01&	SGD	& 0.01	&500	&128\\
    \hhline
  \end{tabular}
\end{table}

We select the widely used traditional TSC algorithm DTW-1NN\cite{15} and the representative deep learning based TSC models FCN\cite{5} and MCNN\cite{6} as comparison. In order to validate the help of SSAM, we also test the base CNN model without SSAM layer.

We visualize loss curve to discuss SSAM's help in accelerating network convergence. Through visulizing the learned mask and SSAM output, we discuss the effectiveness of SSAM. We also validate that L1 regularization can make the model generate a more sparse mask, which plays a role in frequency component selection. At the same time, we validate that SSAM can improve the robustness of the model to noise.

\subsection{Results}
\begin{figure}[!htb]
  \centering
  \includegraphics[width=\hsize]{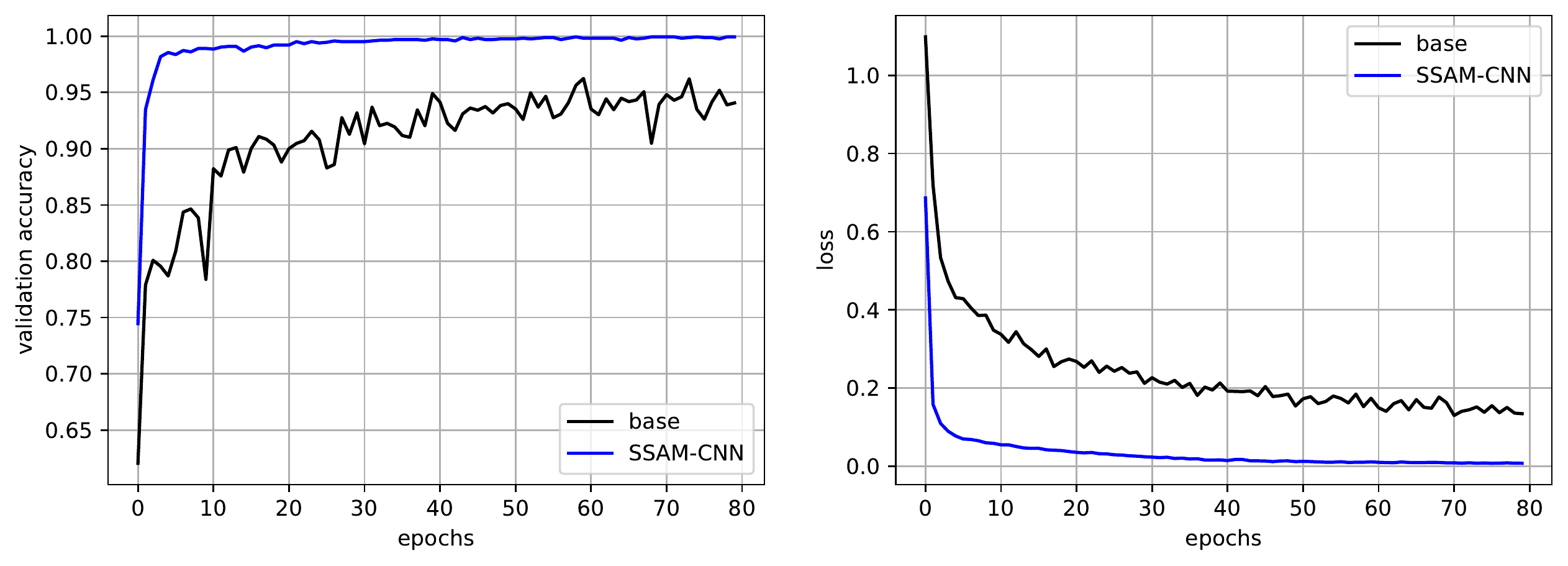}
  \caption{The validation accuracy curve and loss curve on Synthetic dataset.}
  \label{f4}
\end{figure}
Compared to the base model, the introduction of SSAM makes the network converge faster and the loss curve is smoother (see Figure~\ref{f4}). This indicates that SSAM can map the original data into a feature representation that is more conducive to network training.

\begin{figure}[!htb]
  \centering
  \includegraphics[width=\hsize]{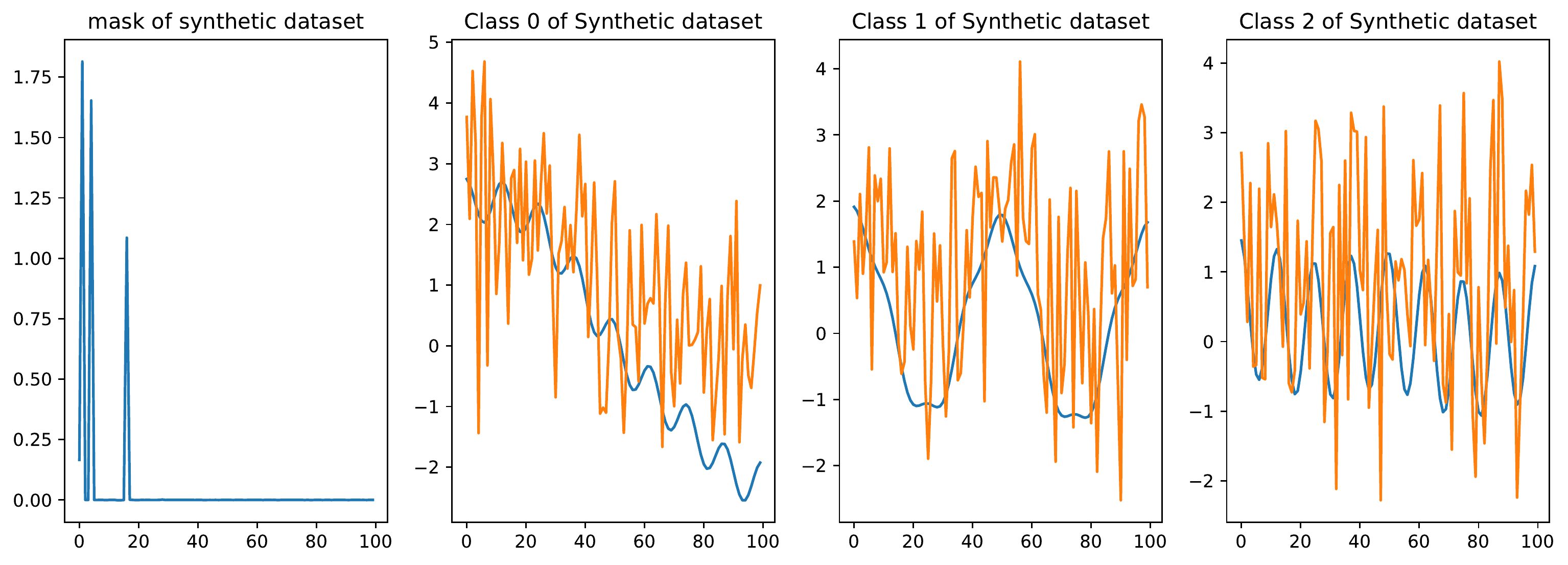}
  \caption{Learned mask and filter output of synthetic dataset.}
  \label{f5}
\end{figure}
The learned mask is sparse and has a larger weight at three frequencies (see Figure~\ref{f5}), which exactly correspond to the three target classes. Therefore, SSAM can indeed assign appropriate weights to each frequency component, highlighting important components and attenuating unimportant components. Due to the influence of noise, the original time series are very similar, more discriminative features are generated by SSAM, making the network easier to train.

\begin{figure}[!htb]
	\centering
	\includegraphics[width=\hsize]{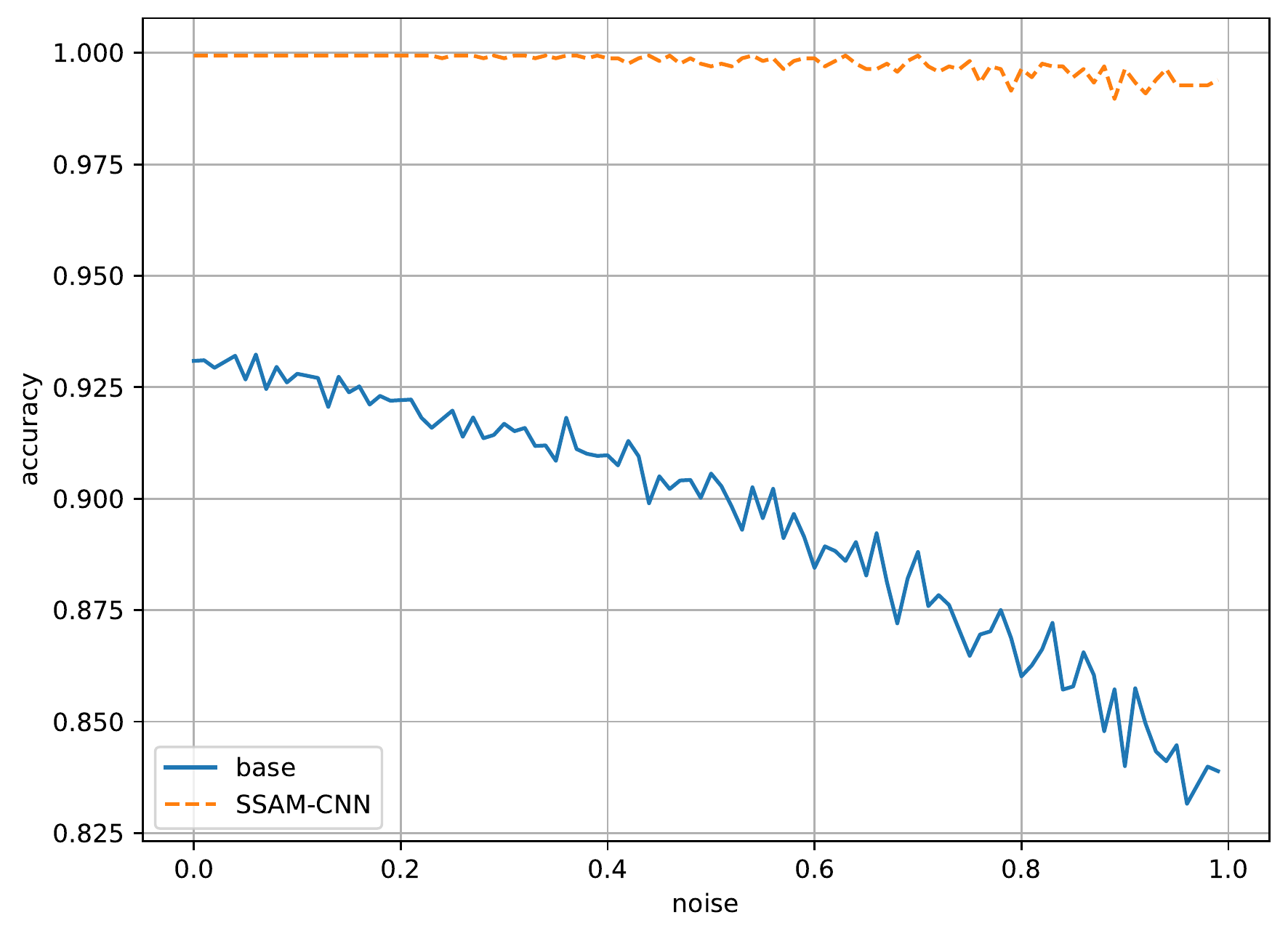}
	\caption{Classification accuracy under different intensity of white noise}
	\label{f6}
\end{figure}
We add white noise to the original data to test the noise immunity of the model. The accuracy of the base model decreases rapidly with the increase of noise intensity, while the accuracy of SSAM-CNN hardly decreases (see Figure~\ref{f6}).  This is because our model is only sensitive to a few frequency components, so it can shield most bands of white noise. Therefore, our model has good robustness.

Table~\ref{t3} shows the test accuracy of all algorithms on each dataset. The hyperparameter $K$ obtained by algorithm3 is marked in the parentheses. From Table~\ref{t3}, the accuracy of SSAM-CNN is higher than all other algorithms on four datasets, and only slightly lower than FCN on CBF dataset. The performance of SSAM-CNN exceeds base model on all datasets, indicating that the introducing SSAM could improve the classification accuracy of the model. The number of segments $K$ obtained by algorithm 3 is distributed between 1 and 3, among which $K$ is 1 on CBF, Face and Synthetic datasets, 2 on Trace datasets, and 3 on CC datasets.
\begin{table}[htbp]
	\centering
	\caption{Comparison results of the algorithms. The best result in each dataset is bolded. $K$ is marked in the parentheses.}
	\label{t3}
	\begin{tabular}{c|c|c|c|c|c}
		\hhline
		Datasets   & \tabincell{c}{DTW\\-1NN} & FCN   & MCNN  & base & \tabincell{c}{SSAM\\-CNN(K)} \\\hline
		CBF   & 91.30 & \textbf{94.60} & 80.43 & 91.30 & 93.48(1) \\\hline
		CC    & 90.33   & 87.78 & 93.33 & 92.22 & \textbf{93.33(3)} \\\hline
		Face  & 82.14 & 91.07 & 89.88 & 83.57 & \textbf{94.64(1)} \\\hline
		Trace & 68.33 & 93.00  & 92.33 & 86.31 & \textbf{93.73(2)} \\\hline
		Synthetic & 63.20 & 88.30 & 93.40 & 93.09 & \textbf{99.94(1)} \\\hhline
	\end{tabular}
\end{table}

\begin{figure}[!htb]
  \centering
  \includegraphics[width=\hsize]{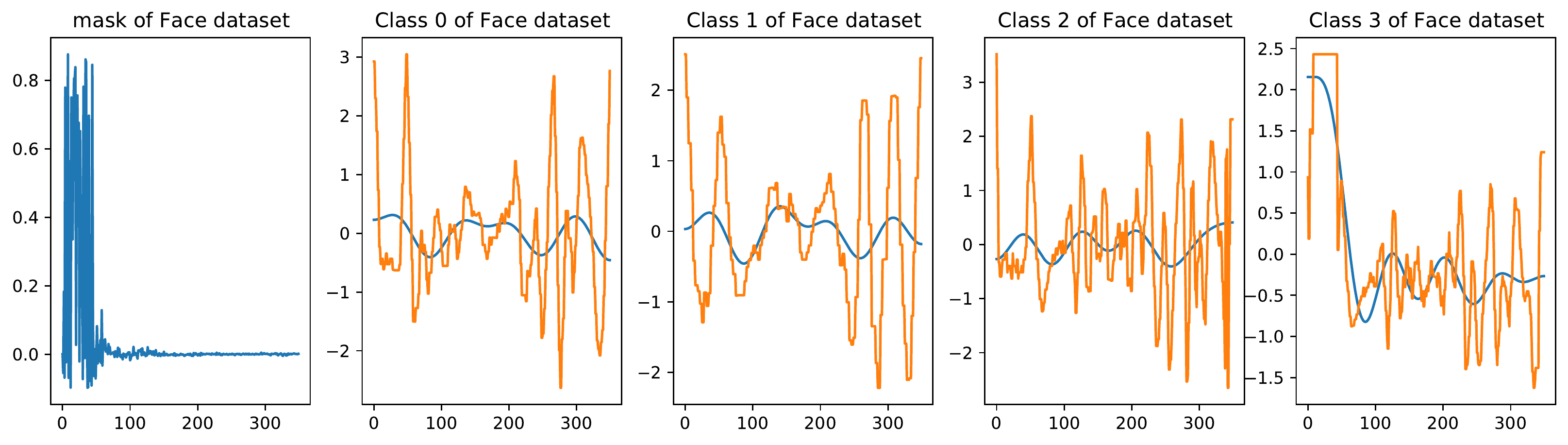}
  \caption{Learned mask and filter output of Face dataset.}
  \label{f7}
\end{figure}
\begin{figure}[!htb]
  \centering
  \includegraphics[width=\hsize]{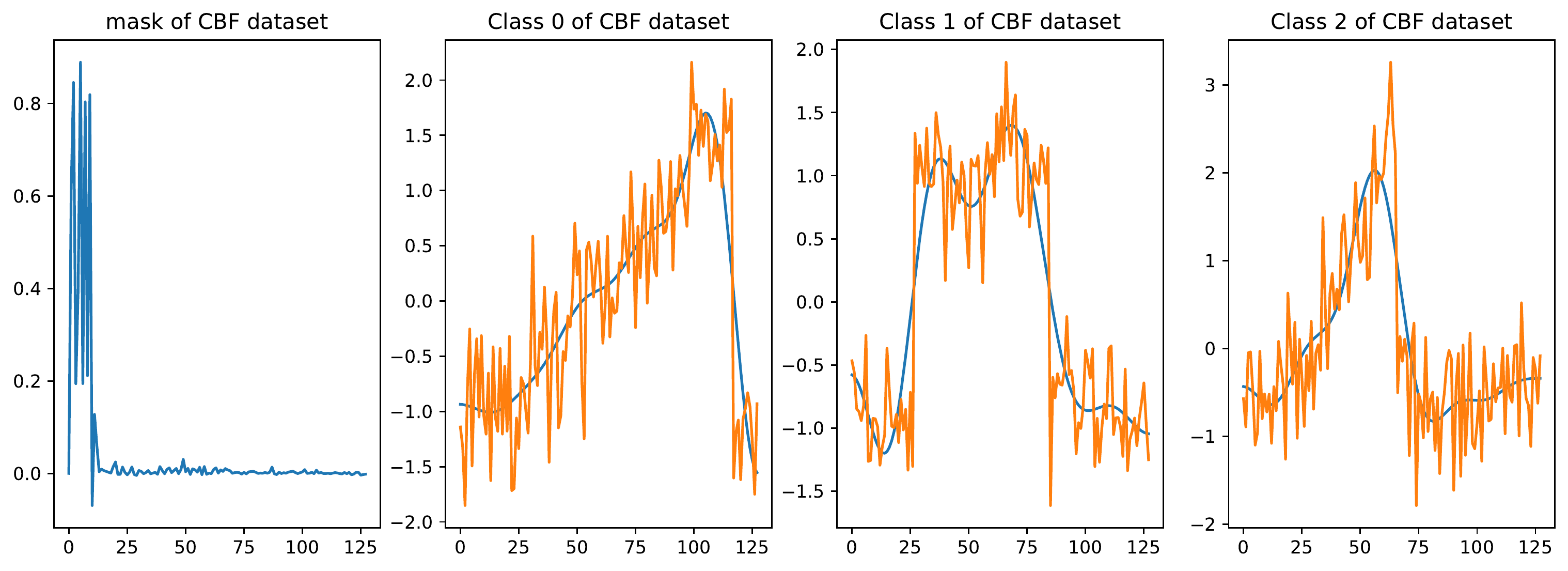}
  \caption{Learned mask and filter output of CBF dataset.}
  \label{f8}
\end{figure}
Figure~\ref{f7} and ~\ref{f8} show the learned mask and filtering results on the real datasets CBF and Face. As can be seen from the figure, SSAM can assign appropriate weights to each frequency component according to the characteristics of data to generate discriminative features. And the learned mask shows that the model pays more attention to the low frequency part of the original data, which is in line with common sense. 

\begin{figure}[!htb]
  \centering
  \includegraphics[width=\hsize]{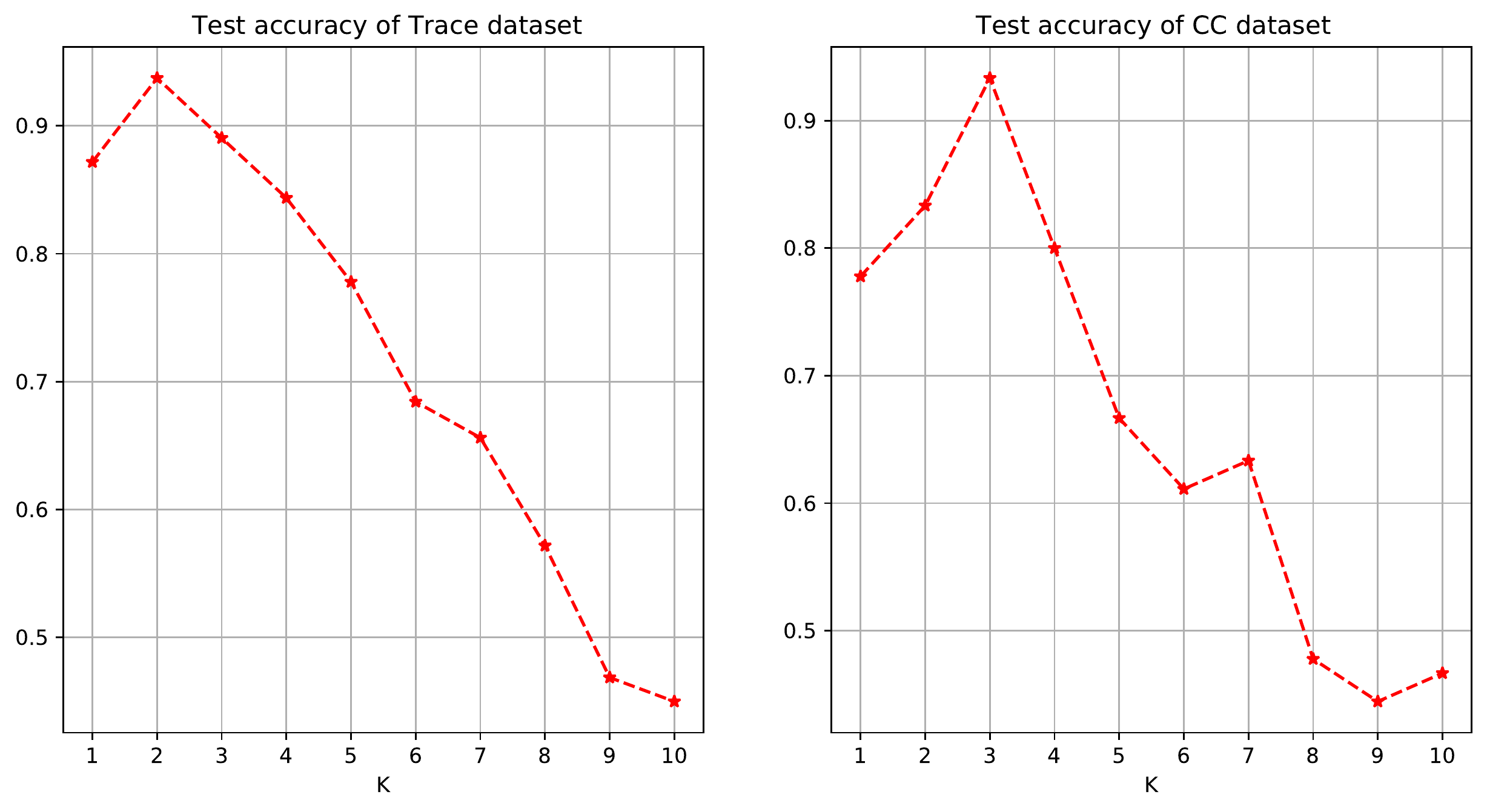}
  \caption{The test accuracy of our model in different segmentation.}
  \label{f9}
\end{figure}
For CC and Trace dataset, the $K$ obtained by Algorithm~\ref{algo3} is not 1. This is because the frequency difference of the original data at different phases may be associated with the label. If the spectrum of the whole sequence is used directly, the phase information will be lost. Therefore, Algorithm~\ref{algo3} gives a $K$ that is not 1, and finally achieves a higher accuracy. In algorithm 3, we use a heuristic strategy to search for $K$, that is, for each candidate value, we only train the model for 5 epochs, and then decide whether to pick the current candidate according to the validation loss. In order to verify the effectiveness of this strategy, we apply each $K$ to the model to obtain the test accuracy. As shown in Fig.~\ref{f9}, when $K$ is 2, the model achieves the highest accuracy on the Trace dataset; When $K$ is 3, the model achieves the highest accuracy on CC dataset. This is the same as the result obtained by the algorithm 3, indicating that the heuristic algorithm can accurately find an appropriate segmentation.

\section{Conclusion}
The main contribution of this paper is to prepose an attention mechanism that acts on the spectrum. In order to avoid the complete loss of time domain information, we also propose a segmented spectral attention mechanism, which uses a tumbling window to segment the original sequence and apply SAM for each segment to preserve the time domain information. We also propose a heuristic algorithm to search for the best number of segments. The experimental results show that the proposed SSAM is able to assign appropriate weights to each frequency components to realize adaptive filtering, which make the model converge faster and smoother, and more robust to noise. And the proposed heuristic search algorithm can indeed find the most suitable segmentation and improve the classification accuracy.

\end{document}